Ari Goodman, Gurpreet Singh, Ryan O'Shea, Peter Teague, James Hing
Naval Air Warfare Center Aircraft Division Lakehurst

# Pose Estimation and Tracking for ASIST

## Abstract

Aircraft Ship Integrated Secure and Traverse (ASIST) is a system designed to arrest helicopters safely and efficiently on ships. Originally, a precision Helicopter Position Sensing Equipment (HPSE) tracked and monitored the position of the helicopter relative to the Rapid Securing Device (RSD). However, using the HPSE component was determined to be infeasible in the transition of the ASIST system due to the hardware installation requirements. As a result, sailors track the position of the helicopters with their eyes with no sensor or artificially intelligent decision aid. Manually tracking the helicopter takes additional time and makes recoveries more difficult, especially at high sea states. Performing recoveries without the decision aid leads to higher uncertainty and cognitive load.

PETA (Pose Estimation and Tracking for ASIST) is a research effort to create a helicopter tracking system prototype without hardware installation requirements for ASIST system operators. PETA leverages advancements in computer vision algorithms to estimate the pose of helicopters in representational camera views.

PETA is meant to improve situational awareness and reduce operator uncertainty with respect to the aircraft's position relative to the RSD, and consequently increase the allowable landing area. The artificial intelligence-based tracking and reporting tool would enable a safer and more efficient recovery process with an increased recovery envelope.

The authors produced a prototype system capable of tracking helicopters with respect to the RSD. The software included a helicopter pose estimation component, camera pose estimation component, and a user interface component. PETA demonstrated the potential for state-of-the-art computer vision algorithms Faster R-CNN and HRNet (High-Resolution Network) to be used to estimate the pose of helicopters in real-time, returning ASIST to its originally intended capability. PETA also demonstrated that traditional methods of encoder-decoders could be used to estimate the orientation of the helicopter and could be used to confirm the output from HRNet.

Future follow-on work will continue when real-world camera feed can be obtained of helicopters using the ASIST system or an analog system.

## Introduction

Aircraft Ship Integrated Secure and Traverse (ASIST) is planned to field in 2024 and originally included a computer vision tracking system. However, the methods used by the original equipment manufacturer (OEM) required aircraft modifications and therefore was not included in the final plans. They had originally planned to include infrared beacons on the helicopters; this approach is robust and could be successful, but it requires modifying the aircraft. The original method was deemed infeasible for every aircraft to receive modifications and recertifications.

PETA is a computer vision-based decision aid and does not require modifying the aircraft. The addition of PETA upon fielding would bring ASIST back to its originally intended capability. PETA would be especially beneficial at high sea states because PETA would drastically reduce the time for the operator to decide if the aircraft is within the Designated Landing Area (DLA). The engineers responsible for determining the safe size of the DLA have stated that the addition of a system of PETA'S nature would allow for increasing the DLA's size, decreasing both the pilot and the ASIST operators' workloads.

In this paper, PETA's software architecture is presented along side its training environment and an evaluation of its performance.

## Background

RGB monocular single-object pose estimation involves estimating the 3D pose of a single object from its 2D RGB image, which can be challenging due to the perspective distortion and occlusion present in real-world images. Although there are numerous pose estimation techniques that become available with more than one camera, the number of cameras available for ASIST operators was initially undetermined [1].

There are several approaches to monocular keypoint-based single-object pose detection, including traditional feature-based methods such as scale-invariant feature transform (SIFT) [2] and speeded up robust features (SURF) [3] which identify distinctive points in an image, as well as more recent deep learning-based methods such as Mask R-CNN [4] and Single Shot Multibox Detector [5] which learn to detect keypoints from data.

Faster R-CNN is one state-of-the-art method in detection and classification [6]. It can be integrated with top-down methods like HRNet by providing a bounding box within which HRNet can detect keypoints [7]. These keypoints can then be interpreted by Perspective-n-Point (PnP) algorithms like EPnP to estimate a final pose [8].

The encoder-decoder network is an alternative architecture which can be used for orientation estimation. In this architecture, the encoder learns a context vector representation of the images, and the decoder learns to output the appropriate rotation value. Encoder-decoders are an older architecture which is better understood and smaller than the aforementioned deep-learning methods. Having multiple agreeing answers from different architectures can be used to increase confidence in the reported pose estimate.

## Methodology

PETA's first stage involved a data collection effort which included searching for existing videos and using subject matter experts (SME) to help construct a realistic simulator. Videos were not available in sufficient quantity, so a simulator was selected as the primary method to generate data. Simulated data can be substituted for real-world data to a degree, but the final system would need to be trained on a real-world dataset [9]. The simulator relied on Unity [10] and included features like accurate deck markings, multiple light levels, helicopter types, and helicopter poses, as seen below in Figure 1. The helicopter and ship models used can be found in [11,12]. Aviation Certification (AVCERT) personnel were contacted to ensure the camera views and helicopter positioning met realism standards. AVCERT personnel's main role is to certify the aircraft facilities, and therefore they have subject matter expertise in what the planned environment will look like.

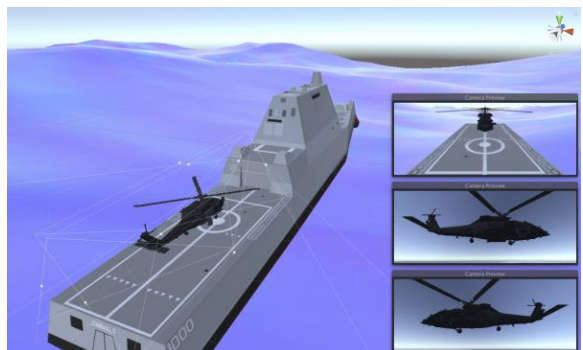

Figure 1: The Simulated Prototype Environment

The next step in the process was data labeling. Data labeling involved associating semantic key points (e.g. nose, cockpit, landing skids, tail rotor, etc.) to pixel coordinates. A skeleton of 19 keypoints was created and used to label every image. An example labeled image can be seen in Figure 2. This process was automated and involved writing software to produce the labels. The simulator was used to create a training dataset of 4,000 images, a testing dataset of 1,000 images, and a final demonstration video.

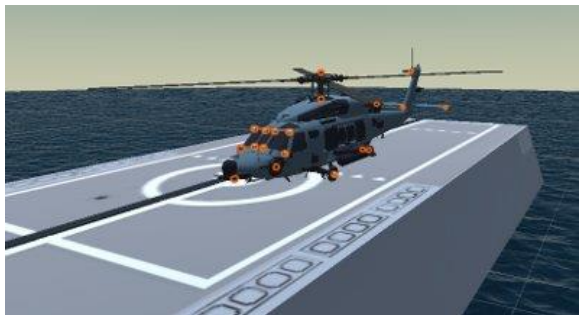

Figure 2: Labeled Keypoints on a Helicopter

The next stage involved training the algorithms on the labeled data. The chosen algorithms were Faster R-CNN, HRNet, and an encoder-decoder network. Faster R-CNN was trained to identify helicopters in the image and to classify the type of helicopter. Faster R-CNN's outputs were passed to HRNet and an encoder-decoder network. HRNet was trained to identify the location of key points, while the encoder-decoder was trained to estimate the rotation of the helicopter. High performance computing resources were leveraged to speed up training. The output from HRNet was a pair of corresponding coordinates in image space and model space, e.g. the nose of the helicopter is located at pixel (X,Y) which was subsequently input into a PnP solving algorithm to estimate the pose.

For Faster R-CNN, the training parameters were: train_batch_size = 1, num_epochs = 10, lr = 0.005, momentum = 0.9, weight_decay = 0.005

For HRNet, the training parameters were: batch_size_per_gpu: 8, shuffle: true, begin_epoch: 0, end_epoch: 120, optimizer: adam, lr: 0.0005, lr_factor: 0.1, lr_step: - 90 - 110 wd: 0.0001, gamma1: 0.99, gamma2: 0.0, momentum: 0.9

The encoder-decoder network utilized a standard convolution autoencoder structure with the addition of a second yaw estimation head. The encoder portion of the network learned to use convolution layers to extract information from the image and distill it down into an encoded representation vector. The decoder portion of the network learned to use inverse convolution layers to reconstruct the input image from the encoded representation vector; the reconstruction loss between images was used as a measure of confidence. Part of the encoded representation vector is also used by the yaw estimation head to calculate the yaw of the object. The yaw estimation head was structured as a fully connected neural network that terminated in overlapping yaw range bins. Each yaw bin (i) represented a set range of rotations that an object could possibly take on. After the most confident bin is selected, the network regressed the final yaw ($\theta$) of the object based on the pre-defined bin centers. An overview of the network can be seen in Figure 3 below.

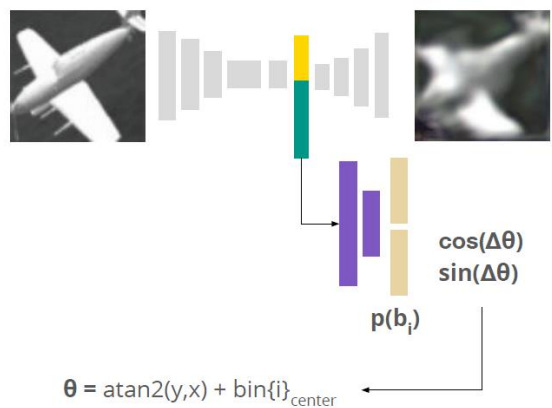

Figure 3: Diagram of Encoder-Decoder Network

Three loss functions were used during training of the network and could be weighted to increase or decrease focus on a specific task. The three loss functions were decoder image reconstruction loss, bin selection loss, and rotational offset loss. The ADAM optimizer was used in conjunction with the three loss functions to train the network for a set number of epochs [13]. Model parameters were saved whenever the model hit a new lowest loss value so the best performing model could be loaded later for testing on other datasets. The authors ran numerous trials to determine the best hyperparameters for the network and found that the following had the greatest effect on network performance: encoded feature vector size, rotational offset loss, portion of the feature vector used for yaw estimation, and rotation bin ranges.

To calculate the final pose error, the sum of the distance between a nominal helicopter pose and the estimated pose was calculate for the X axis, Y axis, and yaw orientation.

In case the future camera would not be calibrated, or vibration would be a significant concern, automatic calibration algorithms were also developed. These algorithms leveraged HRNet to identify paint markings on the deck. These markings have a known real-world coordinate. Knowing the real-world coordinate and pixel coordinate pairs were sufficient to use a direct linear transform (DLT) to estimate changes to external camera parameters.

A user interface was designed to include a visualization of the projection, a measure of confidence, a radial distance measure from the center point of DLA, and a traffic-light style indicator were displayed to the user.

After receiving SME feedback, a new user interface was designed to include clear indication of the status of the helicopter and removed extra information. If the helicopter was determined to be within tolerance near the DLA, then a bounding box was turned green. If the helicopter was determined to not be within tolerance of the DLA, the bounding box was turned red. To ensure users felt confident in the algorithms, the detected key points were also shown. A user could assess if the algorithm had correctly identified the parts of the helicopter, and therefore produced an accurate pose. Values of confidence were deemed not intuitive enough and too distracting to be used in the final decision aid interface.

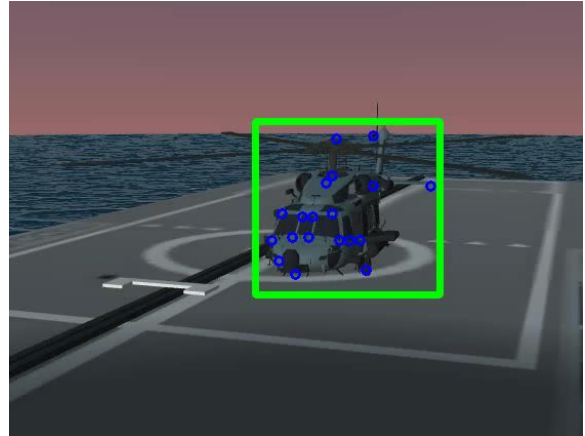

Figure 4: Updated User Interface

## Results

The dataset was presented to AVCERT personnel who approved of using the synthetic environment to evaluate PETA in lieu of real-world data.

The encoder-decoder network was evaluated for rotational accuracy on the test dataset. Below in Figure 5 is a graph of its error and in Figure 6 selected examples of its estimations.

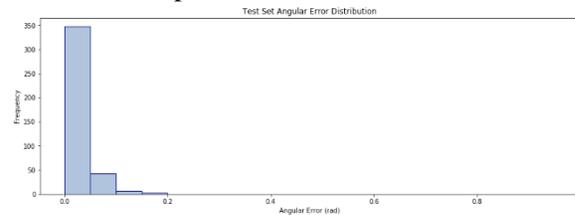

Figure 5: Encoder-Decoder Rotational Error

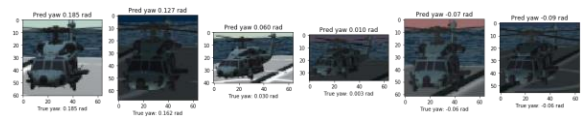

Figure 6: Rotation Estimates from Encoder-Decoder

Its rotational estimates were on average less than 0.1 radian away from optimal in simulation.

A demonstration was given to Air Capable Ships Recovery Team SME's, AVCERT authorities, and PMA 251 Recovery IPT.

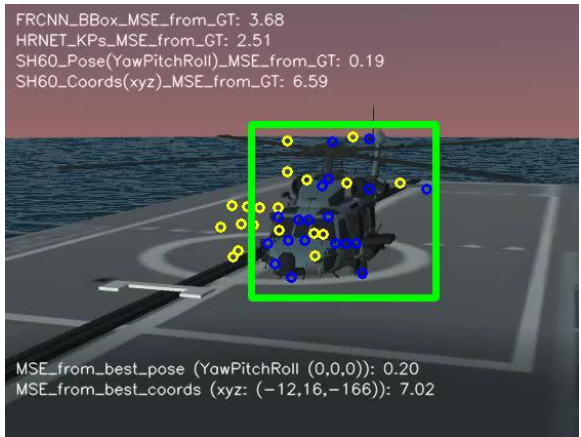

Figure 7: Fully Annotated Demo Video

PETA software that was trained on a subset of this dataset was demonstrated to AVCERT personnel using a custom user-interface. The system including HRNet and Faster R-CNN took on average 90ms per image, or 11 Hz, and dropped to 10Hz with the addition of the decoder and output visualization. PETA was run on 32GB RAM, an i9 processor, and a NVIDIA Quadro P2000 GPU. The system was accurate within 6 inches of the X and Y poses and 0.5 radians in the yaw orientation.

PETA was successful in identifying the deck markers in a variety of lighting environments in synthetic data as shown in Figure 8.

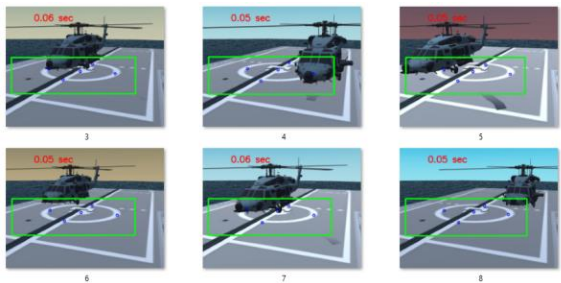

Figure 8: Deck Marking Detection in Various Lighting Conditions

## Conclusion

The purpose of PETA (Pose Estimation and Tracking for ASIST) was to create a helicopter tracking system prototype with simplified reporting interface for the Aircraft Ship Integrated Secure and Traverse (ASIST) System operator. Software requirements mostly focusing on pose estimation accuracy were generated based on AVCERT personnel guidance.

A synthetic environment was developed in Unity to train and evaluate the software and was approved by AVCERT personnel for realism. A prototype computer vision-based approach was developed using Faster R-CNN, HRNet, and an encoder-decoder which could determine the pose of a helicopter relative to the camera position. Once trained, Faster R-CNN accurately and quickly identified the location of the helicopter in the camera view and labeled it with a bounding box. HRNet used the bounding box and image to identify the position of numerous key points, such as the nose of the helicopter. The relative position of the estimated key points to the position of optimal key points was used to determine if the helicopters were in an appropriate position relative to the DLA. The encoder-decoder was used to validate the estimated pose.

The software was shown to have the potential meet requirements without modifying the aircraft. The system was accurate to the nearest half-foot in the X and Y axes and had less than half a radian of error in orientation and operated at 10Hz. The system's working envelope was tested within the required landing regions. The final "in" or "out" of the position decision was output to the user as a green or red rectangle. The algorithms also provided visual ques to the user if it was behaving properly.

PETA performed adequately on synthetic data and is planned to be retrained and retested with real-world footage when made available.

## Acknowledgements

The authors would like to acknowledge Vitaly Ablavsky from the University of Washington for his assistance in designing the encoder-decoder network. The authors would also like to acknowledge Jennifer Spink for interfacing with AVCERT personnel.

Ari Goodman is the S&T AI Lead and a Robotics Engineer in the Robotics and Intelligent Systems Engineering (RISE) lab at Naval Air Warfare Center Aircraft Division (NAWCAD) Lakehurst. In this role he leads efforts in Machine Learning, Computer Vision, and Verification & Validation of Autonomous Systems. He received his MS in Robotics Engineering from Worcester Polytechnic Institute in 2017.

Gurpreet Singh is a Computer Scientist in the Robotics and Intelligent Systems Engineering (RISE) lab at Naval Air Warfare Center Aircraft Division (NAWCAD) Lakehurst. His areas of interest and expertise are Computer Vision and Deep Learning. He received his MS in Computer Science from Stevens Institute of Technology and Graduate Certificate in Robotics Engineering from University of Maryland.

Ryan O'Shea is a Computer Engineer in the Robotics and Intelligent Systems Engineering (RISE) lab at Naval Air Warfare Center Aircraft Division (NAWCAD) Lakehurst. His current work is focused on applying computer vision, machine learning, and robotics to various areas of the fleet in order to augment sailor capabilities and increase overall operational efficiency. He received a Bachelor's Degree in Computer Engineering from Stevens Institute of Technology.

Peter Teague is the Assistant Program Manager for Systems Engineering for the RQ-21A Unmanned Aerial Vehicle. In this role he leads the platform's engineering team and manages system/subsystem requirements and their verification. He previously performed research for Aircraft Launch and Recovery Equipment. Peter received his ME in Mechanical Engineering from Stevens Institute of Technology.

Dr. James Hing serves as the Branch Head of the Strategic Technologies Branch, NAWCAD Lakehurst, where he leads a team of 25 engineers, including 4 R&D Laboratories (IDATS, RISE, INSPIRE, ADHaTEC), in the evaluation, development, maturation, and transition of new and emerging technologies with application to Aircraft Launch Recovery and Support Equipment (ALRE/SE). He has 20 years of multidisciplinary expertise in the fields of computer vision, robotics, and autonomous systems. He received his PhD in Mechanical Engineering from Drexel University.